\documentclass{article}

\usepackage{microtype}
\usepackage{graphicx}
\usepackage{subfigure}
\usepackage{amsmath}
\usepackage{makecell}
\usepackage{multirow}
\usepackage{amsfonts}
\usepackage{bm}
\usepackage{color}
\usepackage{booktabs} 

\usepackage{hyperref}

\usepackage[accepted]{icml2020_revised}

\icmltitlerunning{Deep Quaternion Features for Privacy Protection}

\newcommand{\tabincell}[2]{\begin{tabular}{@{}#1@{}}#2\end{tabular}}

\begin{document}

\twocolumn[
\icmltitle{Deep Quaternion Features for Privacy Protection}



\icmlsetsymbol{equal}{*}

\begin{icmlauthorlist}
\icmlauthor{Hao Zhang}{equal,sjtu}
\icmlauthor{Yiting Chen}{equal,sjtu}
\icmlauthor{Haotian Ma}{southern}
\icmlauthor{Liyao Xiang}{sjtu}
\icmlauthor{Jie Shi}{hua}
\icmlauthor{Quanshi Zhang}{sjtu}
\end{icmlauthorlist}

\icmlaffiliation{sjtu}{Shanghai Jiao Tong University}
\icmlaffiliation{southern}{Southern University of Science and Technology}
\icmlaffiliation{hua}{Huawei International}

\icmlcorrespondingauthor{Quanshi Zhang}{zqs1022@sjtu.edu.cn, John Hopcroft Center and MoE Key Lab of Artificial Intelligence AI Institute, Shanghai Jiao Tong University, China.}

\icmlkeywords{Machine Learning, ICML}

\vskip 0.3in
]



\printAffiliationsAndNotice{\icmlEqualContribution} 

\begin{abstract}
We propose a method to revise the neural network to construct the quaternion-valued neural network (QNN), in order to prevent intermediate-layer features from leaking input information. The QNN uses quaternion-valued features, where each element is a quaternion. The QNN hides input information into a random phase of quaternion-valued features. Even if attackers have obtained network parameters and intermediate-layer features, they cannot extract input information without knowing the target phase. In this way, the QNN can effectively protect the input privacy. Besides, the output accuracy of QNNs only degrades mildly compared to traditional neural networks, and the computational cost is much less than other privacy-preserving methods.

The QNN can be further extended to rotation-equivariant neural networks based on $d$-ary hypercomplex-valued features with better performance~\cite{RENN}.
\end{abstract}

\section{Introduction}
It is often the case that the raw data collected for deep learning cannot be processed locally, so that features are extracted and processed elsewhere. The privacy leakage of those features have been receiving wide attention recently. Many works such as \cite{dosovitskiy2016inverting, zeiler2014visualizing, mahendran2015understanding,
shokri2017membership, Property2018Karan, melis2018exploiting, Yang2019Adversarial}
have pointed out that, attackers can recover significant amount of sensitive input information from the intermediate-layer features of a DNN. As countermeasures, several studies \cite{hybirddeeplearning, PrivyNet, notjustprivacy, PrivacyZhang, SecureML} have been proposed, but have limited applicability mostly due to the exorbitant computational cost.

The task of attribute obfuscation and privacy protection on deep neural networks mainly propose the following two requirements: First, even if attackers have obtained network parameters and
intermediate-layer features, they should not be able to reconstruct the input or infer private attributes of inputs. Second, the privacy-protection method is not supposed to increase the computational cost, or affect the task accuracy significantly.

\begin{figure}[!t]
\begin{center}
\centerline{\includegraphics[width=0.9\columnwidth]{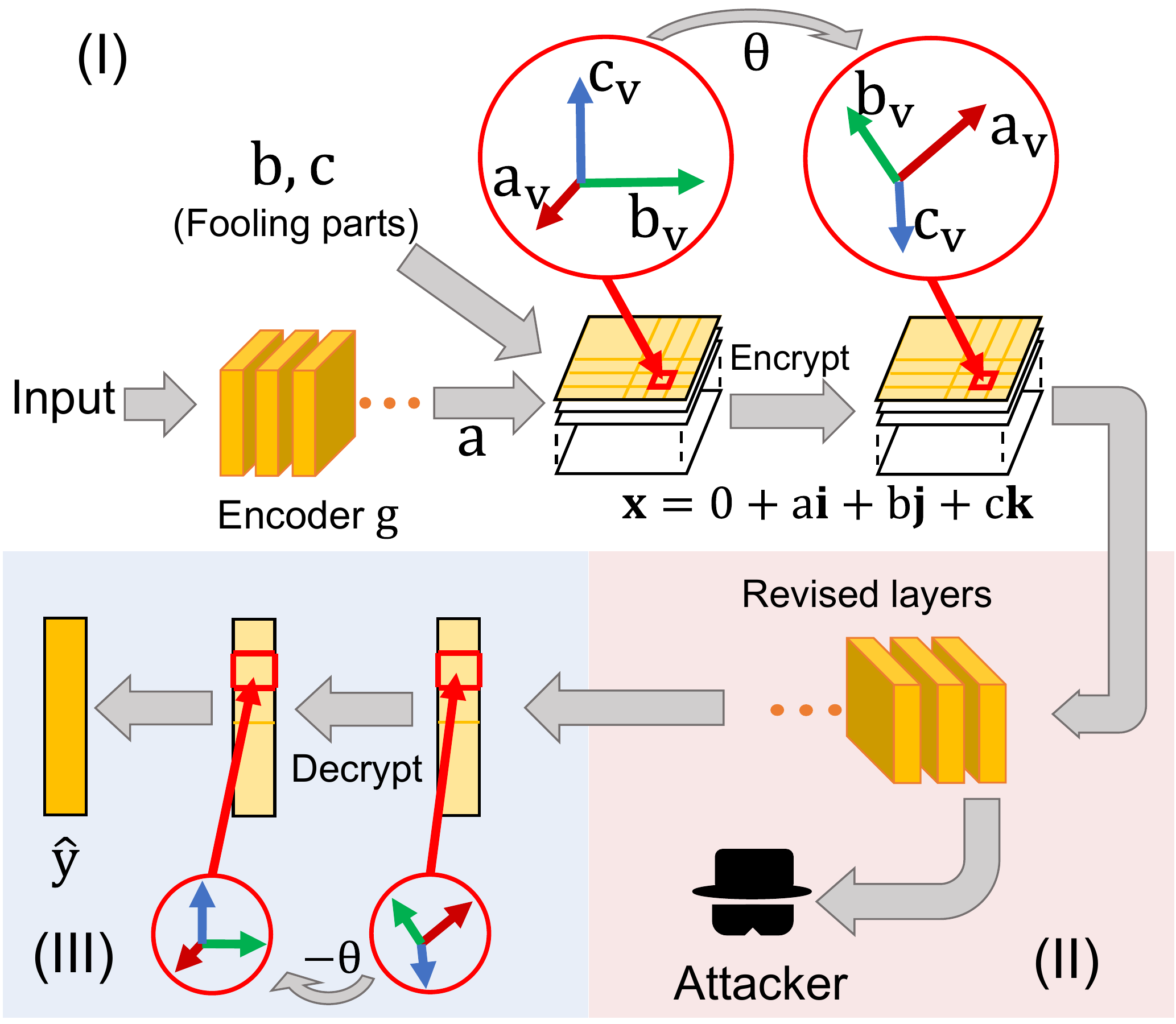}}
\vspace{-5pt}
\caption{Overview of the QNN. The encoder module (I) is located at the local device. The traditional real-valued feature $a$ is extracted and converted into a quaternion-valued tensor for encryption. The quaternion-valued feature is rotated by $\theta$. The rotated feature is processed by the processing module (II) which satisfies the rotation equivariance property. The decoder module (III) uses $\theta$ to decrypt features to get the final result.}
\label{fig:overview}
\end{center}
\vskip -0.2in
\vspace{-10pt}
\end{figure}

Therefore, we propose a generic method to revise a traditional neural network to a quaternion-valued neural network (QNN) to meet the above two requirements. Unlike traditional neural networks, features of the QNN are expressed by quaternion-valued vectors/matrix/tensors. Quaternion is a number system extended from the complex number, which include three imaginary parts and is expressed by $\bm{q}\!=\!q_0\!+\!q_1 \bm{i}\!+\!q_2\bm{j}\!+\!q_3\bm{k}$.

For privacy protection, we hide the input information into a random phase of quaternion-valued features (like the phase of a complex value) in the intermediate-layer of the QNN. The phase can be considered as the private key. Attackers without knowing the target phase cannot recover the input information from quaternion-valued features.

Fig.~\ref{fig:overview} shows the architecture of the QNN. It consists of an encoder, a processing module, and a decoder. The encoder extracts the real-valued feature, converts it into a quaternion-valued feature, and hides the input information in a target phase as the encryption process. The quaternion-valued feature is sent to the processing module for further process. The decoder decrypts the processed feature and computes the result. In order to enable the successful decryption of the decoder, the processing module has to ensure rotation equivariance. \emph{I.e.} if the encoder rotates the target phase containing input information by a certain angle, the phases containing the input information in all layers of the processing module are rotated by the same angle. To achieve that, we need to revise layerwise operations in the neural network, including ReLU, batch-normalization, \emph{etc.} to ensure the rotation equivariance property. However, by exhaustive searching, the attacker is still likely to recover the target phase in quaternion-valued features. Hence we further obfuscate the target phase by adversarial learning. \emph{I.e.} the GAN generates quaternion-valued features to obfuscate the attacker.

The network revision can be broadly applied to DNNs with different architectures for various tasks. Experimental results showed that our QNNs outperformed other baselines in terms of privacy protection, yet the accuracy was not significantly affected.

Our contributions are summarized as follows: We propose a generic method to revise traditional networks into QNNs, which preserves input privacy by hiding sensitive information in a randomly chosen phase. Without knowing the phase, attackers can hardly infer any input information from features. Yet, anyone with the phase information can easily obtain correct outputs without much accuracy loss. Most importantly, QNNs incur far less computational overhead than crypto-based methods.

The QNN can be further extended to rotation-equivariant neural networks based on $d$-ary hypercomplex-valued features with better performance~\cite{RENN}.

\section{Related work}
\textbf{Complex-valued and quaternion-valued neural networks.}
Quaternions are a system extended from traditional complex numbers~\cite{quaternion}.~\cite{Neuronal2014Reichert} applied complex-valued neurons to build richer, versatile representations in neural networks.~\cite{Associative2016Ivo} proposed a method to augment recurrent neural networks with associative memory based on complex-valued vectors, which achieved faster learning on multiple memorization tasks. According to~\cite{Deep2018Trabelsi}, using complex-valued parameters and complex-valued features in neural networks can achieve competitive performance with real-valued networks.

There are also several studies on quaternion-valued networks.~\cite{qcnn} represented color image and convolutional kernels in the quaternion domain to build quaternion convolutional neural networks, which outperformed real-valued neural networks in image classification and denoising tasks.~\cite{qlstm} built quaternion recurrent neural networks and quaternion long-short term memory neural networks, and achieved a better performance than real-valued networks.~\cite{posnet} localized the camera with the help of quaternions. In this paper, we convert intermediate-layer real-valued features of a traditional neural network into quaternion-valued features of a QNN, and hide the input information into a random phase for privacy protection. Unlike~\cite{qcnn, qlstm}, the QNN uses real-valued parameters, instead of quaternion-valued parameters.

\textbf{Privacy protection in deep learning.}
Several studies have been proposed to protect privacy in deep learning.~\cite{hybirddeeplearning} used Siamese fine-tuning to reduce the level of sensitive information in the input, so that attackers cannot infer private properties using intermediate-layer features.~\cite{PrivyNet} proposed the PrivyNet to explore the trade-off between the privacy protection and the task accuracy.~\cite{notjustprivacy} applied a lightweight privacy protection mechanism consisting of data nullification and random noise addition. Homomorphic encryption is a cryptographic technique, which can be applied in the privacy protection in deep learning.~\cite{PrivacyZhang} used the BGV encryption scheme to encrypt data, and used the high-order backpropagation algorithm for training.~\cite{SecureML} distributed data among two non-colluding servers, where the model was trained using secure two-party computation.
~\cite{deepPrivacy} used complex-valued neural networks to protect privacy. Inspired by the homomorphic encryption, we use QNNs. Crucially, compared with the homomorphic encryption, QNNs will not boost the computational cost significantly.

\section{Algorithm}
\subsection{Quaternion}
Quaternion is a number system extended from the complex number. Unlike the complex number, a quaternion consists of three imaginary parts $q_1\bm{i}$, $q_2\bm{j}$, $q_3\bm{k}$, and one real part $q_0$, which is given as $\bm{q}=q_0+q_1 \bm{i}+q_2\bm{j}+q_3\bm{k}$. If the real part of a quaternion is zero ($q_0=0$), we call it a pure quaternion. The quaternion subject to $||\bm{q}|| = \sqrt{q_0^2+q_1^2+q_2^2+q_3^2}=1$ is termed a unit quaternion.
The products of basis elements $\bm{i}$, $\bm{j}$, $\bm{k}$ are given as $\bm{i}^2 = \bm{j}^2 = \bm{k}^2 = \bm{ijk} = -1$, and $\bm{ij}=\bm{k}$,  $\bm{jk}=\bm{i}$,  $\bm{ki}=\bm{j}$, $\bm{ji}=\bm{-k}$, $\bm{kj}=\bm{-i}$, $\bm{ik}=\bm{-j}$.
Note that the multiplication of two imaginary parts is non-commutative, \emph{i.e.} $\bm{ij} \neq \bm{ji}$, $\bm{jk} \neq \bm{kj}$, $\bm{ki} \neq \bm{ik}$. Each quaternion has a polar decomposition. The polar decomposition of a unit quaternion is defined as $\emph{e}^{\bm{o}\frac{\theta}{2}}=\text{cos}\frac{\theta}{2}+\text{sin}\frac{\theta}{2}(o_1\bm{i}+o_2\bm{j}+o_3\bm{k})$, \emph{s.t.} $\sqrt{o_1^2+o_2^2+o_3^2}=1$.

When we use a pure quaternion $\bm{q} = 0+ q_1\bm{i}+ q_2\bm{j}+ q_3\bm{k}$ to represent a point $[q_1,q_2,q_3]^T \in \mathbb{R}^3 $ in a 3D space, the rotation of the point around the axis $\bm{o}=o_1\bm{i}+o_2\bm{j}+o_3\bm{k}$, \emph{s.t.} $\sqrt{o_1^2+o_2^2+o_3^2}=1$, by the angle $\theta$ can be represented as $\bm{Rq\overline{R}}$, where $\bm{R}= \textit{e}^{\bm{o}\frac{\theta}{2}}$, and $\bm{\overline{R}}=\textit{e}^{-\bm{o}\frac{\theta}{2}}$ is the conjugation of $\bm{R}$.

Given a pure quaternion-valued vector $\bm{x}=0+a\bm{i}+b\bm{j}+c\bm{k}\in\mathbb{H}^{n}$, and a real-valued vector $w\in\mathbb{R}^n$, we have

\vspace{-15pt}
\begin{small}
\begin{equation}
\label{eqn:multi}
\bm{x}^Tw=\sum_{v=1}^n \bm{x}_vw_v=0+(a^Tw)\bm{i}+(b^Tw)\bm{j}+(c^Tw)\bm{k}
\end{equation}
\end{small}
\vspace{-15pt}

\subsection{Overview of the QNN}
\label{sec:alg_qnn}
\begin{figure}[t]
\begin{center}
\centerline{\includegraphics[width=0.9\columnwidth]{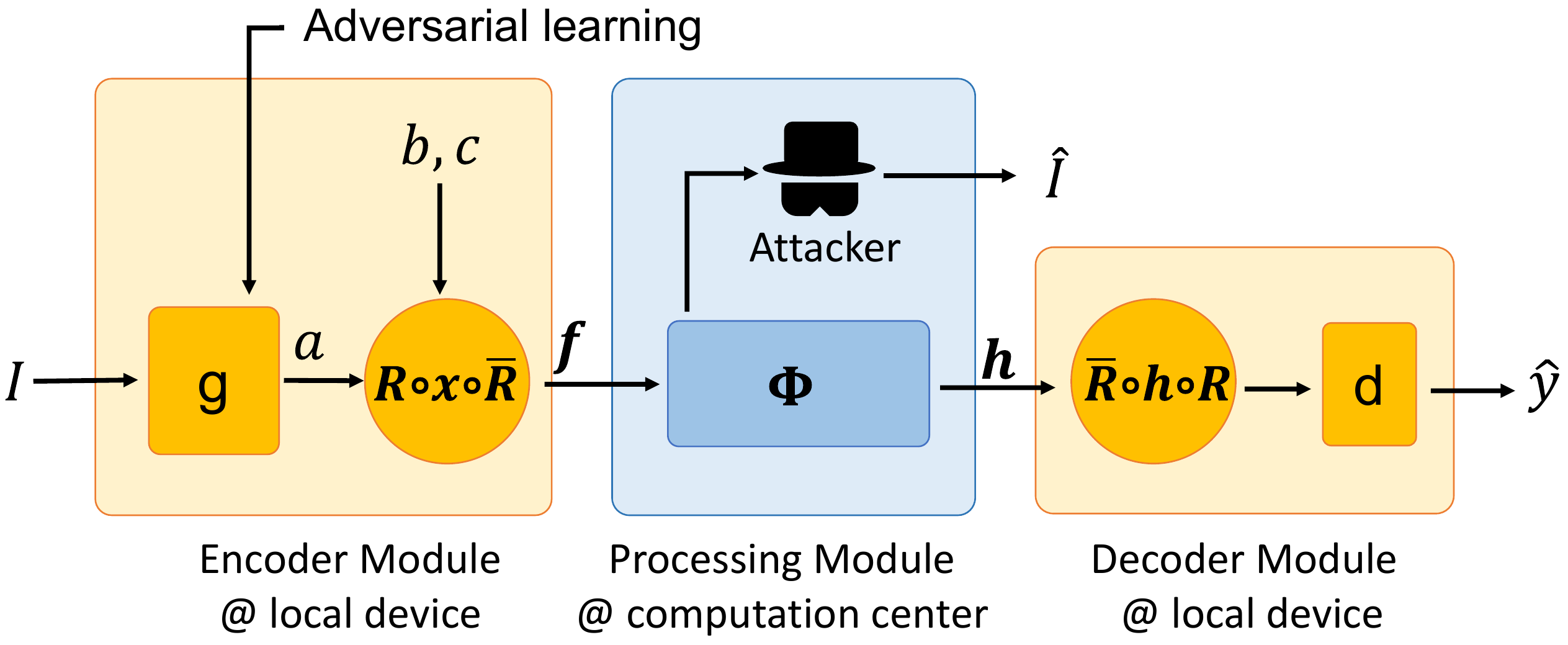}}
\vspace{-10pt}
\caption{Architecture of the QNN. The QNN is divided into three modules: the encoder module, the processing module, and the decoder module. The encoder module extracts real-valued features from the input, and transform it into a quaternion-valued feature. The input information is hidden into a random phase of the quaternion-valued feature. The processing module deals with quaternion-valued features without knowing the target phase. The decoder module decrypts the input information with the target phase to get the final result.}
\label{fig:QNN}
\end{center}
\vspace{-10pt}
\vskip -0.2in
\end{figure}

In this paper, we revise a traditional DNN to construct a QNN. As Fig.~\ref{fig:QNN} shows, we split the QNN into the following three modules:

$\bullet\quad$ \textbf{Encoder module:} The encoder module is usually deployed at the local device. The encoder module extracts traditional real-valued features from the input, and encodes these real-valued features into quaternion-valued features, where each feature element is a quaternion. The encoder uses a specific phase of the quaternion-valued feature to encode the input information, \emph{i.e.} the input information is hidden inside the target phase of each quaternion feature element. Information in other phases of the quaternion feature element represents noises to obfuscate attackers. We consider the target phase as a private key. Without the private key (the target phase), the attacker cannot invert the quaternion-valued back into the input. Then, the encoder module sends the encrypted quaternion-valued feature to the processing module. The target phase is kept by the local device. Note that parameters in the QNN are real-valued, instead of quaternion-valued.

$\bullet\quad$ \textbf{Processing module:} The processing module is supposed to be deployed at the computation center, \emph{e.g.} the cloud, where data can be processed efficiently. The processing module deals with the quaternion-valued feature received from the encoder module. All intermediate-layer features in the processing module are quaternion-valued. The processing module is not given the target phase, which encodes the input information. The task of privacy protection requires the layerwise processing of quaternion-valued features in the processing module to ensure the rotation equivariance property. \emph{I.e.} the input information is hidden inside the same phase of all quaternion-valued features of all layers in the processing module. In this way, the processing module sends the output quaternion-valued feature to the decoder module.

$\bullet\quad$ \textbf{Decoder module:} The decoder module is deployed at the local device. Because of the rotation equivariance of the processing module, the decoder module can use the target phase to decrypt the input information from the received quaternion-valued feature.

\subsection{Detailed design of the QNN}
In this section, we introduce a set of basic rules to transform a traditional neural network into a QNN. We only revise the traditional real-valued feature to the quaternion-valued feature. But parameters in the QNN, \emph{e.g.} weights in a filter, are still real numbers instead of quaternions.

\textbf{Encoder:} Given an input $I\in \textbf{I}$, the encoder module $g$ computes a traditional real-valued feature $a$ as follows.
\begin{small}
\begin{equation}
\centering
\label{eqn:encoder}
a=g(I) \in \mathbb{R}^n
\end{equation}
\end{small}

Then the encoder module uses $a$ and two fooling counterparts $b$, $c$ to generate a quaternion-valued feature $\bm{x} = 0+a\bm{i}+b\bm{j}+c\bm{k}\in \mathbb{H}^n$. Each element in $\bm{x}$ is a quaternion. Note that we can equivalently let $b=g(I)$ or $c=g(I)$ without loss of generality. We encrypt the quaternion-valued feature by rotating $\bm{x}$ along a random axis $\bm{o}=0+o_1\bm{i}+o_2\bm{j}+o_3\bm{k}$ by a random rotation angle $\theta$,
$o_1, o_2, o_3 \in \mathbb{R}$, $||\bm{o}||=1$, and obtain the encrypted feature $\bm{f}\in\mathbb{H}^n$ as follows.
\begin{small}
\begin{equation}
\centering
\label{eqn:encrypt}
\bm{f}=\Psi_{\bm{R}}(a)=\bm{R}\circ\bm{x}\circ\bm{\overline{R}}=\bm{R}\circ(0+a\bm{i}+b\bm{j}+c\bm{k})\circ\bm{\overline{R}}
\end{equation}
\end{small}
where $\Psi_{\bm{R}}(\cdot)$ denotes the function which applies a random rotation $\bm{R}$ to the original quaternion-valued feature $\bm{x}$, $\bm{R}=e^{\bm{o}\frac{\theta}{2}}=\text{cos}\frac{\theta}{2}+\text{sin}\frac{\theta}{2}(o_1\bm{i}+o_2\bm{j}+o_3\bm{k})$, and
$\circ$ denotes the element-wise multipication. The encrypted feature $\bm{f}$ will be sent to the processing module $\Phi$. In this way, we can consider $\bm{Ri}\overline{\bm{R}}$ as the target phase, which encodes the input information, and $\bm{R}$ can be taken as the private key.

\textbf{Processing module:}
Inspired by homomorphic encryption, we revise the operation of each layer in the processing module to satisfy rotation equivariance of the quaternion-valued feature. The rotation equivariance property ensures that the input information is always encoded in the same phase of all quaternion-valued features of all layers in the processing module. In this way, the decoder module can use the target phase to decrypt the input information from the quaternion-valued feature.

The rotation equivariance property can be summarized as follows. If we use $\bm{R}\circ\bm{x}\circ\bm{\overline{R}}$ to rotate quaternion-valued feature $\bm{x}$ along the axis $\bm{o}$ by the angle $\theta$, then quaternion-valued feature elements in each intermediate layer of the processing module are supposed to be rotated along the same axis by the same angle, as follows.
\begin{small}
\begin{equation}
\label{eqn:module_rotate}
\bm{\Phi}(\bm{R}\circ\bm{x}\circ\bm{\overline{R}})=\bm{R}\circ\bm{\Phi}(\bm{x})\circ\bm{\overline{R}}
\end{equation}
\end{small}

Let us consider $\bm{h}_0=\bm{\Phi}(\bm{x})$ without the rotation as the output of the processing module, \emph{i.e.} $\theta=0$, $\bm{R}=e^{\bm{o}\frac{\theta}{2}}=\bm{1}$. The input information is hidden in the imaginary part $\bm{i}$.
Since all parameters in the processing module are real-valued, according to Equation~\eqref{eqn:multi}, the output of the processing module can be represented in the form
\begin{small}
\begin{equation}
\label{eqn:processing1}
\bm{h}_0=\bm{\Phi}(\bm{x})=0+(Aa)\bm{i}+(Ab)\bm{j}+(Ac)\bm{k}\text{.}
\end{equation}
\end{small}
$A$ is a real-valued matrix that represents effects that combine all non-linear transformations in $\bm{\Phi}(\bm{x})$, when $\bm{\Phi}(\bm{x})$ only uses ReLU as non-linear layers. In this way, the input information is still hidden in the imaginary part $\bm{i}$ of $\bm{h}_0$.  Then, let us consider the rotation $\bm{R}$, $\bm{h}=\bm{\Phi}(\bm{R}\circ\bm{x}\circ\overline{\bm{R}})$.
According to Equations~\eqref{eqn:module_rotate} and~\eqref{eqn:processing1}, the output is given as
\begin{small}
\begin{equation}
  \begin{aligned}
  \label{eqn:processing2}
  \bm{h}&=\bm{\Phi}(\bm{R}\circ\bm{x}\circ\overline{\bm{R}})=\bm{R}\circ\bm{h}_0\circ\overline{\bm{R}}\\
  &=(\bm{Ri}\overline{\bm{R}})\circ(Aa)+(\bm{Rj}\overline{\bm{R}})\circ(Ab)+(\bm{Rk}\overline{\bm{R}})\circ(Ac)
\end{aligned}
\end{equation}
\end{small}


In this way, the input information is hidden in the phase $\bm{Ri}\overline{\bm{R}}$ of $\bm{h}$. To ensure the above rotation equivariance, we recursively ensure rotation equivariance of the layerwise operation of each layer inside the processing module. The processing module can be represented as cascaded layers $\Phi(\bm{f})=\Phi_L(\Phi_{L-1}(\cdots\Phi_1(\bm{f}))$, where $\Phi_l(\cdot)$ denotes the $l$-th layer in the processing model. Let $\bm{f}'$ denote the input feature of the $l$-th layer, then the layerwise operation is supposed to satisfy
\begin{small}
\begin{equation}
\label{eqn:single_rotate}
\Phi_{l}(\bm{R}\circ\bm{f}'\circ\bm{\overline{R}})=\bm{R}\circ\Phi_{l}(\bm{f}')\circ\bm{\overline{R}}\text{.}
\end{equation}
\end{small}

Thus, this equation recursively ensures rotation equivariance in Equation~\eqref{eqn:module_rotate}.

In this paper, we revise the operation of each layer to satisfy both Equation~\eqref{eqn:processing2} and~\eqref{eqn:single_rotate}. We consider six most widely-used layers, including the convolutional layer, the ReLU layer, the batch-normalization layer, the avg/max-pooling layer, the dropout layer, and the skip-connection operation.

\emph{Convolutional layer (or fully-connected layer):} For the convolutional layer, we remove the bias term, and get $\text{Conv}(\textbf{f})=w\otimes \textbf{f}$. Note that this revision can also be applied to the fully-connected layer, since the fully-connected layer can be considered as a special convolutional layer.

\emph{ReLU:} We revise the ReLU layer as follows.
\begin{small}
\begin{equation}
\label{eqn:relu}
\text{ReLU}(\bm{f}_v)=\frac{\|\bm{f}_v\|}{\text{max}\{\|\bm{f}_v\|, C\}}\cdot\bm{f}_v
\end{equation}
\end{small}
where $C$ denotes a positive scalar; $\bm{f}_v\in \mathbb{H}$ is the $v$-th element in the quaternion-valued feature $\bm{f}$.

\emph{Batch-normalization:} We revise the batch-normalization operation as follows.
\begin{small}
\begin{equation}
\label{eqn:bn}
\text{norm}(\bm{f}_v^{(k)})=\frac{\bm{f}_v^{(k)}}{\sqrt{\mathbb{E}_{k'}[\|\bm{f}_v^{(k')}\|^2]+\epsilon}}
\end{equation}
\end{small}
where $\bm{f}_v^{(k)}$ denotes the $v$-th quaternion-valued element in the $k$-th sample in the batch. $\epsilon$ denotes a small positive scalar to prevent $\bm{f}_v^{(k)}$ from being divided by zero.

\emph{Avg/Max-pooling:} Avg-pooling layers satisfy Equation~\eqref{eqn:single_rotate} without any revisions. The max-pooling layer selects the quaternion feature element with the largest norm from the receptive field. We revise the max-pooling layer as follows.
\begin{small}
\begin{equation}
\label{eqn:maxpool}
\text{maxpool}(\bm{f})=\bm{f}\circ m,\,\text{where}\, \bm{f}\in \mathbb{H}^{n'}, m\in \{0, 1\}^{n'}
\end{equation}
\end{small}
If the $i$-th element in $\bm{f}$ is selected (the selected quaternion feature element has the largest norm from the receptive field), then $m_i=1$; otherwise, $m_i=0$.

\emph{Dropout:} We randomly dropout several quaternion-valued elements in the feature. The real part and three imaginary parts of dropped elements will all be set to zero.

\emph{Skip connection:} The skip connection can be formulated as $\bm{f}+\Phi(\bm{f})$. If $\Phi(\bm{f})$ satisfies Equation~\eqref{eqn:single_rotate}, then the skip connection will also satisfy Equation~\eqref{eqn:single_rotate}.

All above revised operations satisfy Equation~\eqref{eqn:single_rotate}.


\textbf{Decoder: }
Let $\bm{h}=\bm{\Phi}(\bm{f})$. Let $d$ denote the decoder module, which can be implemented as a shallow network or a simple softmax layer. The decoder module can get the final result $\hat{y}$ as follows.
\begin{small}
\begin{equation}
\label{eqn:decoder}
\hat{y}=d(\Psi_{\bm{R}}^{-1}(\bm{h})),\quad\Psi_{\bm{R}}^{-1}(\bm{h})=\text{Im}_{\bm i}(\overline{\bm{R}}\circ\bm{h}\circ\bm{R})
\end{equation}
\end{small}
where
$\Psi_{\bm{R}}^{-1}(\cdot)$ indicates the inverse function of $\Psi_{\bm{R}}(\cdot)$. The rotation in Equation~\eqref{eqn:decoder} is the inverse of the rotation in Equation~\eqref{eqn:encrypt}. $\text{Im}_{\bm{i}}(\cdot)$ denotes the operation that picks the $\bm{i}$ part from quaternions, and returns a real-valued feature.

\subsection{Encoder based on GAN}
To further boost the robustness to attacks, we need to adopt adversarial learning to learn an encoder. In adversarial learning, we use an encoder to generate quaternion-valued features to fool the attacker from obtaining the target phase. Therefore, we use a GAN to train the encoder.

The GAN~\cite{GAN} includes a generator and a discriminator. In this paper, the generator can be viewed as the encoder module, while the discriminator can be viewed as the attacker. On the one hand, the generator generates quaternion-valued features that satisfy: 1. features contain enough information for the task; 2. features in the target phase and features in the other phase follow the same distribution, so that the discriminator cannot distinguish which phase contains the input information. On the other hand, the discriminator learns to distinguish the target phase that encodes input information.

Given an input $I\in\textbf{I}$ and its real-valued feature $a=g(I)$ from the generator, we can encrpyt it with $\bm{R}$ as $\bm{f}=\Psi_{\bm{R}}(a)=\bm{R}\circ(0+a\bm{i}+b\bm{j}+c\bm{k})\circ\overline{\bm{R}}$, where $b$, $c$ are fooling counterparts that do not contain input information.
The attacker tries to estimate the most probable phase to decrypt $\bm{f}$, and get $a$. Let $\bm{R}'$ denote the rotation estimated by the attacker, and $a'$ denote the decrypted feature, \emph{i.e.} $a'=\Psi_{\bm{R}'}^{-1}(\bm{f})=\text{Im}_{\bm{i}}(\overline{\bm{R}'}\circ\bm{f}\circ\bm{R}')$.
$a$ and $a'$ are inputs of the discriminator $D$, and the discriminator needs to learn to distinguish $a$ and $a'$. The generator $g$ and the discriminator $D$ are trained jointly. We adopt the WGAN~\cite{WGAN} to train the generator and the discriminator as follows.
\begin{small}
\begin{equation}
\label{eqn:wgan}
\begin{aligned}
\quad\min_g &\max_D L(g, D)=\mathbb{E}_{I}[D(a)-\mathbb{E}_{\bm{R}'\neq\bm{R}}[D(a')]]\\
&=\mathbb{E}_{I}[D(a)-\mathbb{E}_{\bm{R}'\neq\bm{R}}[D(\Psi_{\bm{R}'}^{-1}(\Psi_{\bm{R}}(a)))]
\end{aligned}
\end{equation}
\end{small}
To jointly optimize the GAN loss and the task loss, the overall loss is written as follows.
\begin{small}
\begin{equation}
\label{eqn:overall_loss}
\min_{g,\Phi,d}\max_{D} Loss=\min_{g, \Phi, d}[\max_{D}L(g, D)+L_{\text{task}}(\hat{y}, y)]
\end{equation}
\end{small}
where $L_{\text{task}}$ is the loss for the task, and $y$ is the ground-truth label. As for fooling parts $b$ and $c$, we compute them as $b=g(I')$ and $c=g(I'')$ to simplify the implementation, where $I'$ and $I''$ denote images different from $I$.

\subsection{Attackers to QNNs}
\label{sec:alg_attack}
We design two types of attackers to test the performance of the privacy protection of QNNs.

\textbf{Feature inversion attackers:} Let the attacker retrieve quaternion-valued features in the processing module. The feature inversion attacker usually trains another neural network to reconstruct the input.
In this paper, we implement two versions of feature inversion attackers.

$\bullet\quad$ The first inversion attacker aims to estimate the target phase of the quaternion-valued feature, which encodes the input information, \emph{i.e.} to estimate $\bm{R}'$. Then the attacker reconstructs the input using the feature decrypted with $\bm{R}'$, $a'=\Psi_{\bm{R}'}^{-1}(\bm{f})$. The attacker learns another neural network to reconstruct the input, \emph{i.e.} $\hat{I}=\text{dec}_1(a')$.

Thus, the core technique is to enumerate all phases, and train a classifier $D'$ to judge whether the decrypted feature $a'$ encodes the input information. The classifier $D'$ is similar to the discriminator $D$ in the GAN.

$\bullet\quad$ The second feature inversion attacker directly uses the encrypted quaternion-valued feature to reconstruct the input. The attacker is trained and tested with $\bm{f}=\Psi_{\bm{R}}(a)$, \emph{i.e.} $\hat{I}=\text{dec}_2(\bm{f})=\text{dec}_2(\Psi_{\bm{R}}(a))$.

\textbf{Property inference attackers:} According to~\cite{Property2018Karan}, the ``property'' refers to the input attribute. We design the first property inference attacker to use the intermediate-layer feature to infer sensitive attributes of inputs. For example, if we train a neural network for the age recognition task, then the property such as gender and race should also be hidden in intermediate-layer features. In this paper, we test four versions of property inference attackers.

$\bullet\quad$ The first property inference attacker trains a classifier, which uses the quaternion-valued feature to infer sensitive attributes of the input. Specifically, the attacker decrypts $\bm{f}$ using an estimated $\bm{R}'$ as $a'=\Psi_{\bm{R}'}^{-1}(\bm{f})$. The classifier $D'$ is used to judge whether the decrypted feature $a'$ encodes the input information. The attacker learns another classifier to estimate $attr=\text{net}(I)$. The attacker trains the classifier using pairs of inputs and their attributes, $(I, attr)$, and tests the classifier using the reconstructed input $\hat{I}$.

The property inference attacker is supposed to deal with quaternion-valued features of any layer in the processing module. However, all intermediate-layer features are computed based on the input feature $\bm{f}$, which contains the richest information of the input. To simplify the implementation, we only consider hacking the feature $\bm{f}$ in this paper. This is uniformly applied to the 2nd, 3rd, and 4th attackers.

$\bullet\quad$ The second property inference attacker is similar to the first property inference attacker. The only difference is that the attacker trains the classifier using pairs of the decrypted feature $a'$ and ground-truth attributes, $(a', attr)$. \emph{I.e.} the attacker learns the classifier $attr=\text{net}(a')$, and is tested using the obtained $a'$ during the attack process.

$\bullet\quad$ The third property inference attacker is similar to the first property inference attacker. The attacker learns the classifier using annotations of reconstructed inputs and attributes, $(\hat{I}, attr)$, \emph{i.e.} $attr=\text{net}(\hat{I})$. The classifier is tested using the reconstructed inputs $\hat{I}$.

$\bullet\quad$ The forth property inference attacker applies the \emph{k}-nearest neighbors (\emph{k}-NNs) instead of a neural network to infer sensitive attributes. For images whose decrypted features $a'$ are close to each other in the feature space, we consider these images have similar attributes. In this way, the attacker uses the $k$-nearest neighbors in the feature space to infer the sensitive attributes.

\subsection{K-anonymity}
Interestingly, we can analyze the performance of the QNN from the perspective of the $k$-anonymity. \emph{K-anonymity} means that when attackers try to reconstruct an input from the quaternion-valued feature, they can get at least $k$ different results, while only one of them is the correct input. We visualize the decrypted result $\hat{I}$ obtained from the feature inversion attacker or the property inference attacker for analysis. Fig.~\ref{fig:k-anonymity} shows reconstructed inputs from different phases with similar confidence estimated by the attacker. Table~\ref{tab:k-anonymity} shows that the QNN can roughly achieve the $k$-anonymity.

\section{Experiments}

In this section, we conducted a series of experiments to test  QNNs. We converted a number of classic neural networks into QNNs. Theoretically, the QNN can be applied to different tasks. In this paper, we tested QNNs on tasks include object classification and face attribute estimation, considering both the feature inversion attacker and the property inference attacker.

We trained QNNs using the CIFAR-10, CIFAR-100 datasets~\cite{cifar}, which contained small images, for object classification. Besides, we used the CUB200-2011 dataset~\cite{CUB200} and the CelebA dataset~\cite{CelebA} for object classification and face attributes estimation with large image. We revised classic neural networks into QNNs, which included LeNet~\cite{lenet}, residual networks~\cite{resnet}, VGG-16~\cite{vgg}, and AlexNet~\cite{alexnet}.

\begin{figure}[t]
\begin{center}
\centerline{\includegraphics[width=0.9\columnwidth]{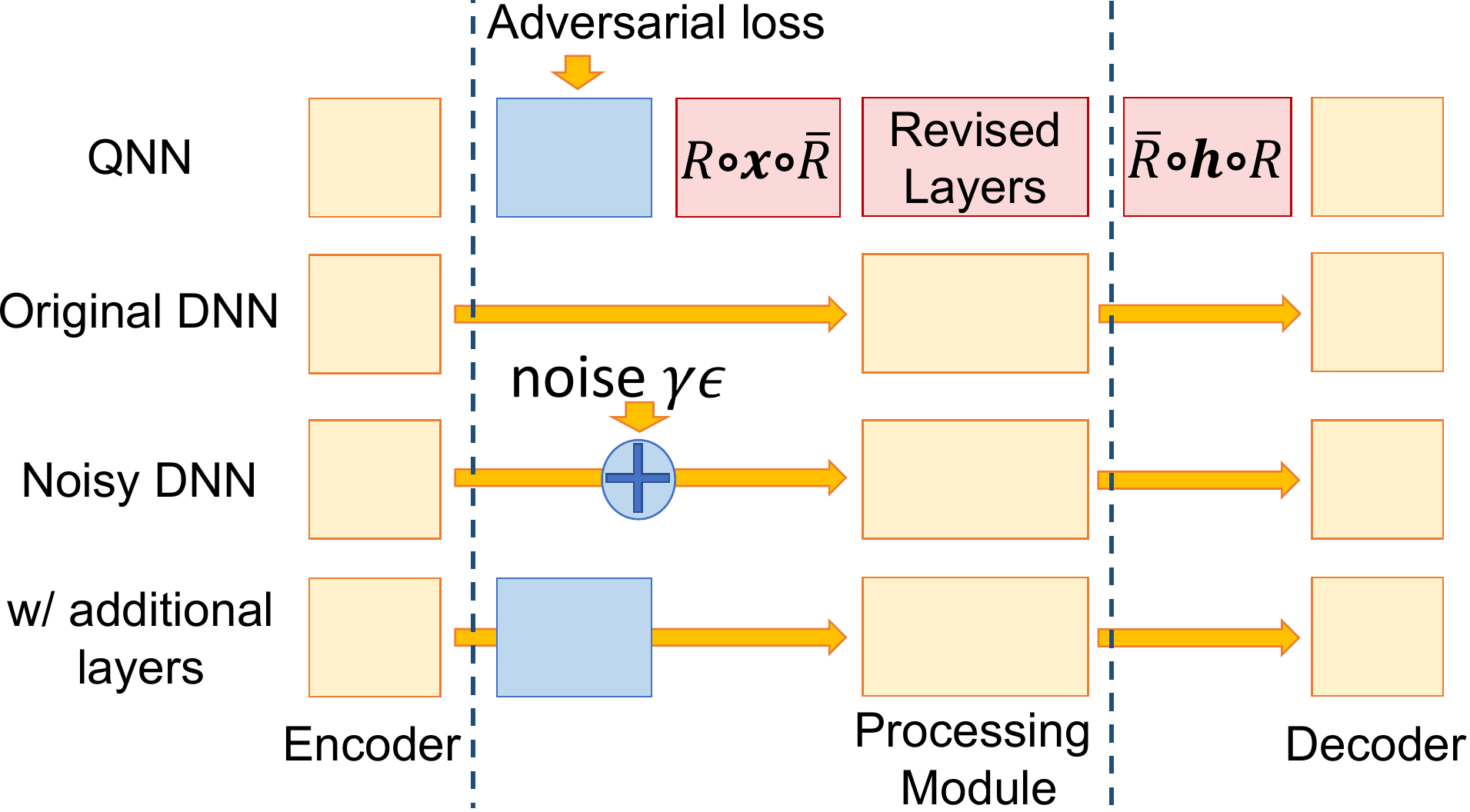}}
\vspace{-5pt}
\caption{Alignment of architectures between neural networks.}
\label{fig:structure}
\end{center}
\vskip -0.2in
\vspace{-10pt}
\end{figure}

\begin{table*}[!t]
\caption{Classification error rates and reconstruction errors indicating the capacity of privacy protection.}
\label{tab:resnet_accuracy}
\begin{center}
\resizebox{0.85\linewidth}{!}{\
\begin{tabular}{c|cc|cccc|c|cccccc}
\toprule
\multirow{6}*{\rotatebox{90}{\tabincell{c}{Classification\\ Error Rate(\%)}}} \!\!\!\!&\!\!\!\! Model &\!\!\!\!\!\! Dataset &\!\!\!\! \tabincell{c}{Original\\DNN} \!\!\!\!&\!\!\!\! \tabincell{c}{w/ additional\\ layers} \!\!\!\!&\!\!\!\! \tabincell{c}{Complex\\NN} \!\!\!\!&\!\!\!\! \tabincell{c}{QNN}
\!\!\!\!&\!\!\!\! \multirow{6}*{\rotatebox{90}{\tabincell{c}{Reconstruction\\ Errors}}} \!\!\!\!&\!\!\!\!\tabincell{c}{Original\\DNN} \!\!\!\!&\!\!\!\! \tabincell{c}{w/ additional\\ layers} \!\!\!\!&\!\!\!\! \tabincell{c}{Complex\\ dec($a'$)} \!\!\!\!&\!\!\!\! \tabincell{c}{Complex\\ dec($x$)} \!\!\!\!&\!\!\!\! \tabincell{c}{Quaternion\\ dec($a'$)} \!\!\!\!&\!\!\!\! \tabincell{c}{Quaternion\\ dec($x$)} \\
\cmidrule{2-7}
\cmidrule{9-14}
 &
ResNet-20 &\!\!\!\!\!\! CIFAR-10 & 11.56 & 9.68 & 10.91 & \textbf{9.21} & & 0.0906 & 0.1225 & 0.2664 & 0.2420 & \textbf{0.3014} & 0.2702\\
&ResNet-32 &\!\!\!\!\!\! CIFAR-10 & 11.13 & 9.67 & 10.48 & \textbf{9.82} & & 0.0930 & 0.1171 & 0.2569 &
0.2412 & \textbf{0.2813} & 0.2412 \\
&ResNet-44 &\!\!\!\!\!\! CIFAR-10 & 10.67 & 9.43 & 11.08 & \textbf{9.54} & & 0.0933 & 0.1109 & 0.2746 &
0.2419 & \textbf{0.3123} & 0.2421 \\
&ResNet-56 &\!\!\!\!\!\! CIFAR-10 & 10.17 & 9.16 & 11.53 & \textbf{9.24} & & 0.0989 & 0.1304 & 0.2804 & 0.2377 & \textbf{0.3083} & 0.2403 \\
&ResNet-110 &\!\!\!\!\!\! CIFAR-10 & 10.19 & 9.14 & 11.97 & \textbf{9.31} & & 0.0896 & 0.1079 & \textbf{0.3081} &
0.2495 & 0.3028 & 0.2379 \\
\bottomrule
\end{tabular}
}
\resizebox{0.85\linewidth}{!}{\
\begin{tabular}{c|cc|cc|ccc|cccc}
\toprule
\multirow{7}*{\rotatebox{90}{\tabincell{c}{Reconstruction\\ Errors}}} \!\!\!\!&\!\!\!\! Model \!\!\!\!&\!\!\!\! Dataset \!\!\!\!&\!\!\!\! \tabincell{c}{Original\\DNN} \!\!\!\!&\!\!\!\! \tabincell{c}{w/additional\\ layers} \!\!\!\!&\!\!\!\! \tabincell{c}{Noisy DNN\\$\gamma=0.2$} \!\!\!\!&\!\!\!\! \tabincell{c}{Noisy DNN\\$\gamma=0.5$}
 \!\!\!\!&\!\!\!\! \tabincell{c}{Noisy DNN\\$\gamma=1.0$} \!\!\!\!&\!\!\!\! \tabincell{c}{Complex\\dec($a'$)} \!\!\!\!&\!\!\!\!
\tabincell{c}{Complex\\dec($x$)} \!\!\!\!&\!\!\!\! \tabincell{c}{Quaternion\\dec($a'$)} \!\!\!\!&\!\!\!\!
\tabincell{c}{Quaternion\\dec($x$)} \\
\cmidrule{2-12}
& LeNet&\!\!\!\! CIFAR-10 \!\!\!\!& 0.0769 & 0.1208 & 0.0948 & 0.1076 & 0.1274 & 0.2405 & 0.2353 & \textbf{0.2877} & 0.2303 \\
& LeNet&\!\!\!\! CIFAR-100 \!\!\!\!& 0.0708 & 0.1314 & 0.0950 & 0.1012 & 0.1286 & 0.2700 & 0.2483 & \textbf{0.2996} & 0.2528 \\
& ResNet-56 &\!\!\!\! CIFAR-100 \!\!\!\!& 0.0929 & 0.1029 & 0.1461 & 0.1691 & 0.2017 & 0.2593 & 0.2473 & \textbf{0.3057} & 0.2592 \\
& ResNet-110 &\!\!\!\! CIFAR-100 \!\!\!\!& 0.1050 & 0.1092 & 0.1483 & 0.1690 & 0.2116 & 0.2602 & 0.2419 & \textbf{0.3019} & 0.2543\\
& VGG-16 &\!\!\!\! CUB200-2011 \!\!\!\!& 0.1285 & 0.1202 & 0.1764 & 0.0972 & 0.1990 & 0.2803 & 0.2100 & \textbf{0.3133} & 0.1945 \\
& AlexNet &\!\!\!\! CelebA \!\!\!\!& 0.0687 & 0.1068 & - & - & - & \textbf{0.3272} & 0.2597 & 0.3239 & 0.2657 \\

\bottomrule
\end{tabular}
}
\end{center}
\vskip -0.1in
\vspace{-5pt}
\end{table*}

\begin{table}[!t]
\caption{Classification error rates on variety of models and datasets.}
\label{tab:other_accuracy}
\vspace{-8pt}
\begin{center}
  \resizebox{\linewidth}{!}{\
  \begin{tabular}{lc|cc|ccc|cc}
  \toprule
  Model \!\!\!\!&\!\!\!\! Dataset \!\!\!\!&\!\!\!\! \tabincell{c}{Original\\DNN} \!\!\!\!&\!\!\!\! \tabincell{c}{w/ additional\\ layers} \!\!\!\!&\!\!\!\! \tabincell{c}{Noisy DNN\\$\gamma=0.2$} \!\!\!\!&\!\!\!\! \tabincell{c}{Noisy DNN\\$\gamma=0.5$}
   \!\!\!\!&\!\!\!\! \tabincell{c}{Noisy DNN\\$\gamma=1.0$} \!\!\!\!&\!\!\!\! \tabincell{c}{Complex\\NN} \!\!\!&\!\!\! QNN \\
  \midrule
  LeNet & CIFAR-10 & 19.78 & 21.52 & 24.15 & 27.53 & 34.43 & 17.95 \!\!\!&\!\!\! \textbf{11.45} \\
  LeNet& CIFAR-100 & 51.45 & 49.85 & 56.65 & 67.66 & 78.82 & 49.76 \!\!\!&\!\!\! \textbf{37.78}\\
  ResNet-56 & CIFAR-100 & 53.26 & 44.38 & 57.24 & 61.31 & 74.17 & \textbf{44.37} \!\!\!&\!\!\! 44.86 \\
  ResNet-110 & CIFAR-100 & 50.64 & 44.93 & 55.19 & 61.12 & 71.31 & 50.94 \!\!\!&\!\!\! \textbf{42.05} \\
  VGG-16 \!\!\!\!\!&\!\!\!\!\!\! CUB200-2011 \!\!\!\!\!\!&\!\!\!\! 56.78 & 63.47 & 69.20 & 99.48 & 99.48 & 78.50 \!\!\!&\!\!\! \textbf{70.86} \\
  AlexNet & CelebA & 14.17 & 9.49 & - & - & - & 15.94 \!\!\!&\!\!\! \textbf{8.80}\\
  \bottomrule
  \end{tabular}
  }
\end{center}
\vskip -0.1in
\end{table}

\begin{table*}[!t]
\caption{Failure rate of identifying the reconstructed image by human annotators.}
\label{tab:error_human}
\begin{center}
\resizebox{0.85\linewidth}{!}{\
\begin{small}
\begin{tabular}{lc|cc|ccc|cccc}
\toprule

Model \!\!\!\!&\!\!\!\! Dataset &\!\!\!\! \tabincell{c}{Original\\DNN} \!\!\!\!&\!\!\!\! \tabincell{c}{w/ additional\\ layers} \!\!\!\!&\!\!\!\! \tabincell{c}{Noisy DNN\\$\gamma=0.2$} \!\!\!\!&\!\!\!\! \tabincell{c}{Noisy DNN\\$\gamma=0.5$}
 \!\!\!\!&\!\!\!\! \tabincell{c}{Noisy DNN\\$\gamma=1.0$} \!\!\!\!&\!\!\!\! \tabincell{c}{Complex\\dec($a'$)} \!\!\!\!&\!\!\!\!
\tabincell{c}{Complex\\dec($x$)} \!\!\!\!&\!\!\!\! \tabincell{c}{Quaternion\\dec($a'$)} \!\!\!\!&\!\!\!\!
\tabincell{c}{Quaternion\\dec($x$)} \\
\midrule
LeNet & CIFAR-10 & 0.16 & 0.12 & 0.20 & 0.20 & 0.24 & 0.82 & 0.92 & 0.90 & \textbf{0.96} \\
LeNet & CIFAR-100 & 0.16 & 0.12 & 0.20 & 0.64 & 0.72 & 0.80 & 0.92 & 0.94 & \textbf{1.00}\\
ResNet-56 & CIFAR-100 & 0.06 & 0.06 & 0.08 & 0.10 & 0.36 & 0.72 & 0.88 & 0.90 & \textbf{1.00} \\
ResNEt-110 & CIFAR-100 & 0.04 & 0.12 & 0.10 & 0.16 & 0.36 & 0.80 & 0.86 & 0.94 & \textbf{0.98} \\
VGG-16 \!\!\!\!&\!\!\!\! CUB200-2011 & 0.06 & 0.06 & 0.08 & 0.02 & 0.14 & 0.86 & 0.84 & \textbf{0.86} & 0.74\\
AlexNet & CelebA & 0.04 & 0.24 & - & - & - & 0.96 & 1.00 & 0.84 & \textbf{1.00} \\
\bottomrule
\end{tabular}
\end{small}
}
\end{center}
\vspace{-5pt}
\vskip -0.1in
\end{table*}

\textbf{Network architectures:}
Fig.~\ref{fig:structure} compares architectures between the original neural network and its corresponding QNN.
The encoder/processing/decoder modules of each original neural network are introduced as follows. For the LeNet, the encoder consisted of all layers before the second convolutional layer, and there was only a softmax-layer in the decoder. For the residual network, the encoder consisted of all layers before the first $16\times16$ feature map. The decoder consisted of all layers after the first $8\times8$ feature map. For the AlexNet, the encoder consisted of the first convolutional layers, and the decoder consisted of fully-connected layers and the softmax layer. For the VGG-16, all layers before the last $56\times 56$ feature map compromised the encoder, and the decoder consisted of fully-connected layers and the softmax layer. Note that for each encoder, we add the GAN at the end of the encoder.

\textbf{Baselines:}
We proposed four baselines for comparison.
As shown in the second row of Fig.~\ref{fig:structure}, the original network without any revision was used as the first baseline which was denoted as \emph{Original DNN}. We divided the original network into encoder, processing module, and decoder in the same way as QNN.
The third row of Fig.~\ref{fig:structure} shows the second baseline. Noise addition could also be used for privacy protection. We added noises to the output $a$ of the encoder, \emph{i.e.} $a+\gamma\epsilon$, where $\epsilon$ denoted a random noise vector, and $\gamma$ was a scalar.
The second baseline was denoted as \emph{Noisy DNN}, and we trained \emph{Noisy DNN} with $\gamma=0.2,0.5,1.0$. The third baseline is shown in the last row of Fig.~\ref{fig:structure}, which was denoted as \emph{w/ additional layers}. Since the insertion of the GAN increased the layer number of the QNN. For a fair comparison, we also added the GAN architecture into the baseline network, but the baseline network was learned without the GAN loss.
The forth baseline was the ~\cite{deepPrivacy}, which was constructed using the same division of modules of the QNN. For simplicity, the forth baseline was denoted as \emph{Complex NN}. Theoretically speaking, QNNs had more potential phases to hide the input information than Complex NNs, and used more fooling counterparts to obfuscate attackers. Besides, the rotation in 3D space was more complicated than rotation in 2D space. Therefore, the QNNs were supposed to have a better privacy protection performance than Complex NNs.

\begin{figure*}[t]
\begin{center}
\centerline{\includegraphics[width=0.75\linewidth]{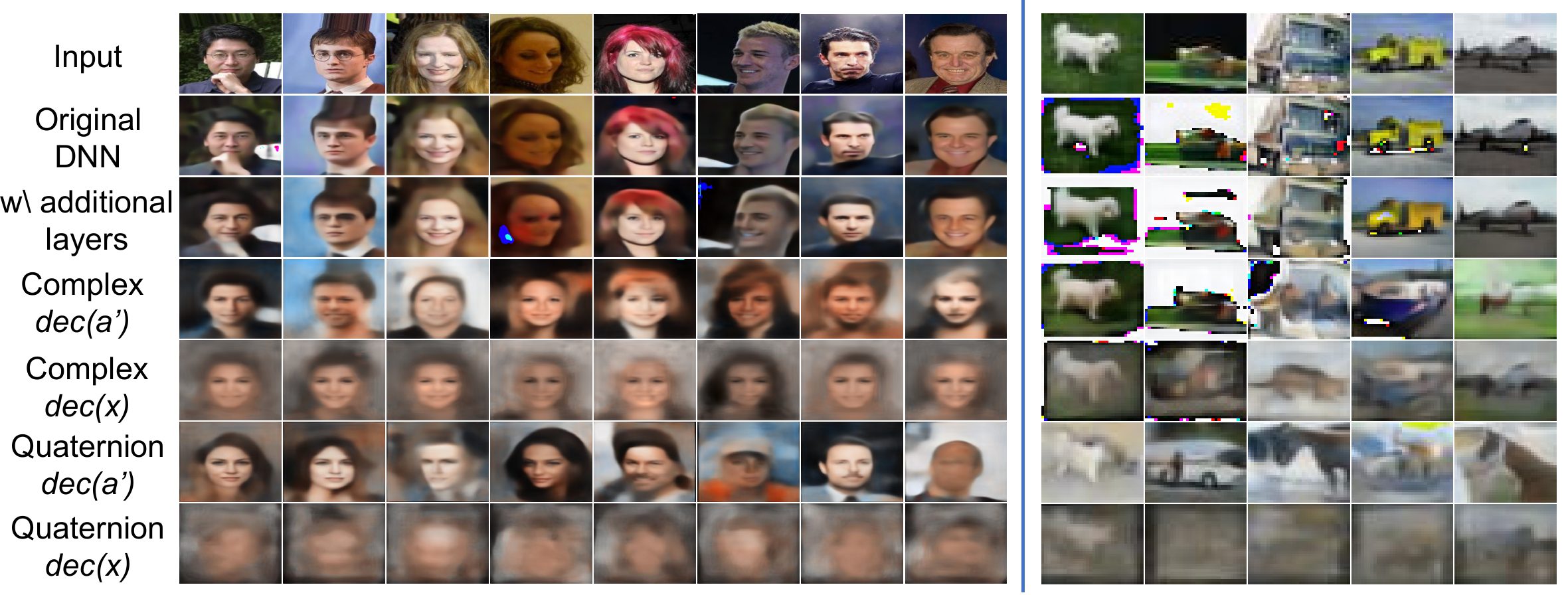}}
\vspace{-5pt}
\caption{Reconstructed examples. Left: images from the CelebA dataset; right: images from the CIFAR-10 dataset.}
\label{fig:reconstruct}
\end{center}
\vskip -0.2in
\vspace{-10pt}
\end{figure*}

\textbf{Attack:}
We applied two kinds of attackers in Sec.~\ref{sec:alg_attack}.

$\bullet\,$ \emph{Inversion Attack:}
The inversion attacker was implemented based on U-Net~\cite{UNet}.We revised the U-Net to construct the attacker for inversion attacks. The intermediate-layer feature was upsampled to the size of the input, which was fed into the inversion model. There were four down-sample blocks and four up-sample blocks. Each block had six convolutional layers for better performance. The output of the inversion model had the same size as the input. We used intermediate-layer features from QNNs as the input of the attacker. In this way, we used the U-Net to construct both the first inversion attacker and the second inversion attacker, according to Sec.~\ref{sec:alg_attack}. The first inversion attacker trained the attacker with real-valued features, and tested the attacker with the output of $D$. The second inversion attack trained and tested the attacker with the encrypted quaternion-valued features.
To mimic the procedure of hacking the privacy, we randomly sampled the rotation axis and rotation angle for 1000 times. The sample with the highest output of the discriminator was considered as the optimal feature to reconstruct the original image.

$\bullet\,$ \emph{Inference attacker:}
For the inference attack, we used the CelebA and CIFAR-100 datasets for testing. For the CelebA dataset, we selected 10 attributes as private attributes. We revised an AlexNet to a QNN, and trained the QNN to estimate other 30 attributes, whereas the attacker based on ResNet-50 used the intermediate-layer feature to estimate private attributes. For the CIFAR-100 dataset, we trained a quaternion-valued ResNet-56 to classify major 20 superclasses of CIFAR-100. The attacker was constructed based on ResNet-56, and used the intermediate-layer feature to infer 100 minor classes, which were considered as sensitive information in this experiment.

\begin{table}[t]
\caption{Average error of the estimation angle and the failure rate. Complex error and quaternion error represented the average error of estimated phase on Complex NN and QNN, respectively.}
\label{tab:error_angle}
\vspace{-10pt}
\begin{center}
\resizebox{\linewidth}{!}{\
\begin{small}
\begin{tabular}{lc|cc}
\toprule
& & \multicolumn{2}{c}{Average Error} \\
\midrule
Model & Dataset & \tabincell{c}{Complex\\ Error} & \tabincell{c}{Quaternion\\ Error}\\
\midrule
ResNet-20 & CIFAR-10 & 0.7890$\pm$0.3722 & 1.4803$\pm$0.7004\\
ResNet-32 & CIFAR-10 & 0.7820$\pm$0.3630 & 1.3322$\pm$0.6785\\
ResNet-44 & CIFAR-10 & 0.8411$\pm$0.6200 & 1.4610$\pm$0.6905\\
ResNet-56 & CIFAR-10 & 0.8088$\pm$0.5848 & 1.4733$\pm$0.6932\\
ResNet-110 & CIFAR-10 & 0.8048$\pm$0.4535 & 1.4461$\pm$0.6955\\
LeNet & CIFAR-10 & 0.7884$\pm$0.4147 & 1.3511$\pm$0.6765 \\
LeNet & CIFAR-100 & 0.8046$\pm$0.5279 & 1.3842$\pm$0.6694 \\
ResNet-56 & CIFAR-100 & 0.7898$\pm$0.5544 & 1.3837$\pm$0.7010\\
ResNet-110 & CIFAR-100 & 0.7878$\pm$0.3775 & 1.3515$\pm$0.6558\\
VGG-16 \!\!\!\!\!\!\!\!&\!\!\!\!\!\!\!\! CUB200 \!\!\!\!\!\!\!\!\!\!&\!\!\!\! 1.5572$\pm$0.8778 & 1.3589$\pm$0.7120 \\
AlexNet & CelebA & 0.8500$\pm$0.5811 & 1.3833$\pm$0.6767 \\
\bottomrule
\end{tabular}
\end{small}
}
\end{center}
\vskip -0.1in
\end{table}

\begin{table}[t]
\caption{Experimental results of inference attackers. net($I$), net($a'$), net($\hat{I}$), $k$-NN represented the first, the second, the third, and the forth inference attacker, respectively.}
\label{tab:inference_error}
\vspace{-5pt}
\begin{center}
\resizebox{\columnwidth}{!}{\
\begin{small}
\begin{tabular}{c|c|c|c|c|c|ccc}
\toprule
& &\!\!\!\! \tabincell{c}{Classification\\Error Rate} \!\!\!\!&\!\!\!\!
net($I$) \!\!\!\!&\!\!\!\! net($a'$) \!\!\!\!&\!\!\!\! net($\hat{I}$) \!\!\!\!&  \multicolumn{3}{c}{$k$-NN} \\
\midrule
 & Structure & & & & &\!\!\!\! $k=1$ \!\!\!\!&\!\!\!\! $k=3$ \!\!\!\!&\!\!\!\! $k=5$ \!\!\!\!\\
\midrule
\multirow{3}*{\rotatebox{90}{\tabincell{c}{CIFAR-\\100}}} \!\!\!\!\!\!&\!\!\!\! w/ additional layers \!\!\!\!& 18.73 & 68.72 & 38.50 & 40.28 & 73.38 & 68.37 & 71.16\\
 & Complex NN & 26.77 & 94.53 & 87.17 & 89.56 & 94.44 & 93.63 & 92.50 \\
 & QNN & 21.60 & \textbf{96.73} & \textbf{94.22} & \textbf{95.25} & \textbf{98.19} & \textbf{97.60} & \textbf{97.42}  \\
\midrule
\multirow{3}*{\rotatebox{90}{CelebA}} \!\!\!\!\!\!&\!\!\!\! w/ additional layers \!\!\!\!& 8.04 & 19.14 & 13.17 & 14.01 & 20.14 & 17.26 & 16.20 \\
 & Complex NN & 14.75 & \textbf{25.72} & \textbf{22.21} & 22.61 & 31.69 & 27.90 & 26.41 \\
 & QNN & 8.20 & 25.03 & 22.19 & \textbf{23.26} & \textbf{32.77} & \textbf{28.81} & \textbf{27.42}\\
\bottomrule
\end{tabular}
\end{small}
}
%
\end{center}
\vskip -0.1in
\vspace{-5pt}
\end{table}

\begin{table}[t]
\caption{Rank of the estimated sample (first row), and time cost of inference (second row).}
\label{tab:k-anonymity}
\vspace{-5pt}
\begin{center}
\resizebox{0.8\linewidth}{!}{\
\begin{small}
\begin{tabular}{c|cccc}
\toprule
\!\!\!\!\!\!&\!\!\!\! \tabincell{c}{w/ additional\\ layers} \!\!\!\!&\!\!\!\!\tabincell{c}{Complex\\NN} \!\!\!\!&\!\!\!\! QNN \!\!\!\!&\!\!\!\! \tabincell{c}{Homomorphic\\encryption} \\
\midrule
Rank \!\!\!\!& - \!\!\!\!&\!\!\!\!\!\!\!\! 36 \!\!\!\!&\!\!\!\!\!\! 525.6 \!\!\!\!\!\!&\!\!\!\! -\\
\tabincell{c}{Time cost\\ (s/image)} \!\!\!\!& 0.0004 \!\!\!\!&\!\!\!\!\!\!\!\! 0.0007 \!\!\!\!&\!\!\!\! 0.0011 \!\!\!\!&\!\!\!\! 3.56 \\
\bottomrule
\end{tabular}
\end{small}
}
\end{center}
\vskip -0.1in
\vspace{-5pt}
\end{table}

\begin{figure}[t]
\begin{center}
\centerline{\includegraphics[width=0.8\columnwidth, height=0.5\columnwidth]{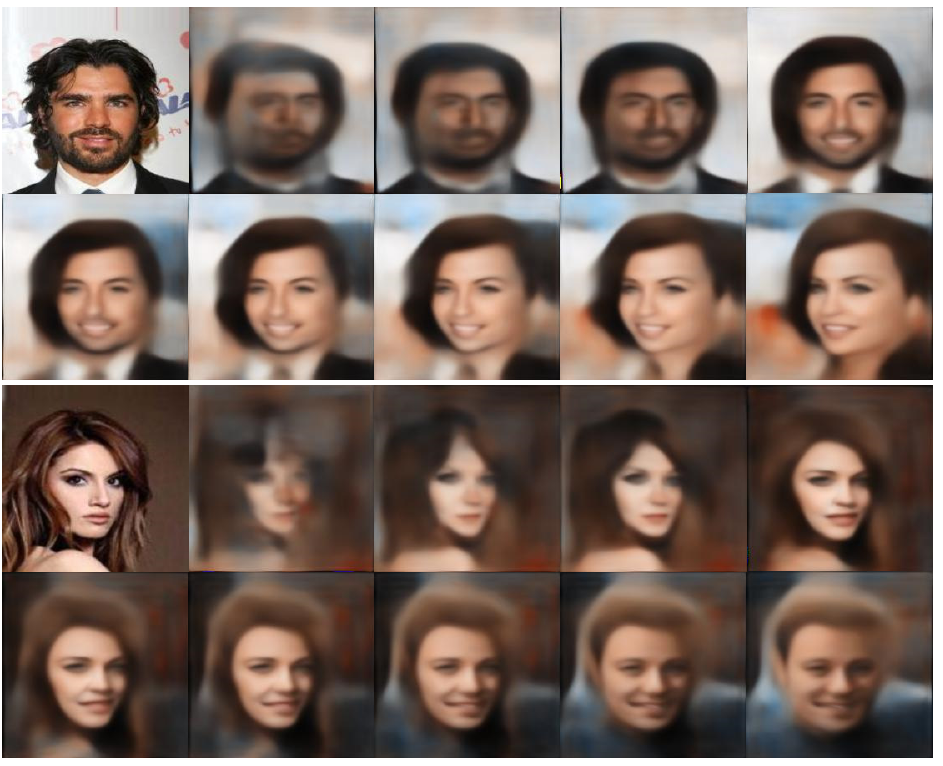}}
\vspace{-5pt}
\caption{CelebA images reconstructed using different phases. The first image of every two rows is the input image. We pick the most meaningful images.}
\label{fig:k-anonymity}
\end{center}
\vskip -0.2in
\vspace{-5pt}
\end{figure}

\textbf{Evaluation metrics of privacy protection:}
We used four metrics to evaluate the performance of privacy protection in terms of inversion attackers: (1) the pixel-level reconstruction error $\mathbb{E}[\|\hat{I} - I\|]$, where the pixel value was scaled to $[0, 1]$. (2) The average difference angle $\Delta \theta$ between $\textbf{x}^{R^*}$ and $\textbf{x}^{\hat{R}}$, where  $\textbf{x}^{R^*}=\bm{R^*i}\overline{\bm{R}}^*$
and $\textbf{x}^{\hat{R}}$
were unit quaternion vectors of \textbf{x}-axis after rotation using $R^*$ and $\hat{R}$, respectively. (3) The rank of the estimated sample. Let us recover two samples using two similar phases. According to our experience, when the angle between phases was less than $\Delta\theta=5^\circ$, the two samples usually presented the same object entity. Thus, we estimated the number of object entities that were more similar to the input than the recovered sample. For the QNN, the rank of the estimated sample is computed as $\frac{2}{1-\text{cos}\Delta\theta}$. For the Complex NN, the rank of the estimated sample is computed as $\frac{2\pi}{\Delta\theta}$. (4) The reconstruction failure rate of human identification, \emph{i.e.} we used human annotators to judge whether they can identify the input based on the reconstructed sample $\hat{I}$. For the privacy performance of inference attacks, we could use the accuracy of attackers as the evaluation metric. What's more, we can compute the processing speed of different models to measure the efficiency of QNNs.


\textbf{Experimental Results and Analysis:}
Table~\ref{tab:resnet_accuracy} and Table~\ref{tab:other_accuracy} show the performance of QNNs and baselines. Complex $dec(a')$ and Complex $dec(x)$ represented the first inversion attacker and the second inversion attacker on Complex NN, respectively. Quaternion $dec(a')$ and Quaternion $dec(x)$ represented the first inversion attacker and the second inversion attacker on QNN, respectively. Classification error rate results showed that QNNs achieved a better performance than Complex NNs. Compared with Original DNN and DNNs with additional layers, the accuracy of QNNs was not significantly affected. As for the reconstruction error, a higher value indicated a better privacy protection performance. The reconstruction error of QNNs was higher than other networks, \emph{i.e.} QNNs exhibited better performance of privacy protection than other networks.

Fig.~\ref{fig:reconstruct} visualizes several examples from inversion attackers. We only provided results from partial experiments constrained by the space. Table~\ref{tab:error_angle} shows averages and standard errors of $\Delta \theta$. A smaller value of the average of $\Delta \theta$ indicated that attackers were easier to estimate the target phase. Table~\ref{tab:error_human} shows the subjective failure rate according to the judgement of humans. Attackers could not fetch the input information from encrypted features. It was more difficult to estimate the target phase of quaternion-valued features than complex-valued features.


Table~\ref{tab:inference_error} shows the result of inference attackers. A higher value of inference error indicated a better privacy protection performance. Attackers on QNNs had a higher inference classification error than attackers on DNNs with additional layers, and attackers on Complex NN. Thus, QNNs protected private attributes from attackers more effectively.

Fig.~\ref{fig:k-anonymity} shows images reconstructed using different phases. Table~\ref{tab:k-anonymity} shows the rank of the estimated sample. A higher rank value indicated better performance of privacy protection. The rank of the QNN was higher than the Complex NN. \emph{I.e.} it was more difficult for attackers on QNNs to find the target phase than attackers on Complex NNs.

Table~\ref{tab:k-anonymity} also shows the time cost of the inference process of DNNs. The time cost of homomorphic encryption was from the framework Gazelle~\cite{Gazelle}, which trained a small network with 3 fully-connected layers from~\cite{SecureML} using the CIFAR-10 dataset. For other networks, we used networks revised from the ResNet-56, which were deeper than the network used by Gazelle. However, the inference time cost of the QNN was much less than Gazelle, and was comparable with traditional DNNs.

\section{Conclusion}
In this paper, we propose a method to protect the privacy of inputs. Our method transforms traditional neural networks into QNNs, which use quaternion-valued features as intermediate-layer features. The input information is hidden in a random phase of quaternion-valued features. Experiments showed the effectiveness of the privacy protection of QNNs. The QNN has much lower computational cost than the homomorphic encryption.

\bibliography{privacy}

\begin{thebibliography}{32}
\providecommand{\natexlab}[1]{#1}
\providecommand{\url}[1]{\texttt{#1}}
\expandafter\ifx\csname urlstyle\endcsname\relax
  \providecommand{\doi}[1]{doi: #1}\else
  \providecommand{\doi}{doi: \begingroup \urlstyle{rm}\Url}\fi

\bibitem[Arjovsky et~al.(2017)Arjovsky, Chintala, and Bottou]{WGAN}
Arjovsky, M., Chintala, S., and Bottou, L.
\newblock Wasserstein generative adversrial networks.
\newblock \emph{In Proceedings of the 34th International Conference on Machine
  Learning (ICML)}, pp.\  214--223, 2017.

\bibitem[Danihelka et~al.(2016)Danihelka, Wayne, Uria, Kalchbrenner, and
  Graves]{Associative2016Ivo}
Danihelka, I., Wayne, G., Uria, B., Kalchbrenner, N., and Graves, A.
\newblock Associative long short-term memory.
\newblock \emph{{In} Proceedings of the 33rd International Conference on
  Machine Learning (ICML)}, 2016.

\bibitem[Dosovitskiy \& Brox(2016)Dosovitskiy and
  Brox]{dosovitskiy2016inverting}
Dosovitskiy, A. and Brox, T.
\newblock Inverting visual representations with convolutional networks.
\newblock \emph{Proceedings of the IEEE Conference on Computer Vision and
  Pattern Recognition (CVPR)}, pp.\  4829--4837, 2016.

\bibitem[Ganju et~al.(2018)Ganju, Wang, Yang, Gunter, and
  Borisov]{Property2018Karan}
Ganju, K., Wang, Q., Yang, W., Gunter, C.~A., and Borisov, N.
\newblock Property inference attacks on fully connected neural networks using
  permutation invariant representations.
\newblock In \emph{Proceedings of the 2018 ACM SIGSAC Conference on Computer
  and Communications Security}, pp.\  619--633, 2018.

\bibitem[Goodfellow et~al.(2014)Goodfellow, Pouget-Abadie, Mirza, Xu,
  Warde-Farley, Ozair, Courville, and Bengio]{GAN}
Goodfellow, I., Pouget-Abadie, J., Mirza, M., Xu, B., Warde-Farley, D., Ozair,
  S., Courville, A., and Bengio, Y.
\newblock Generative adversarial nets.
\newblock \emph{In {NIPS}}, 2014.

\bibitem[Hamilton(1848)]{quaternion}
Hamilton, W.~R.
\newblock On quaternions; or on a new system of imaginaries in algebra.
\newblock \emph{The London, Edinburgh and Dublin Philosophical Magazine and
  Journal of Science}, 33\penalty0 (219):\penalty0 58--60, 1848.

\bibitem[He et~al.(2016)He, Zhang, Ren, and Sun]{resnet}
He, K., Zhang, X., Ren, S., and Sun, J.
\newblock Deep residual learning for image recognition.
\newblock \emph{In {CVPR}}, 2016.

\bibitem[Juvekar et~al.(2018)Juvekar, Vaikuntanathan, and
  Chandrakasan]{Gazelle}
Juvekar, C., Vaikuntanathan, V., and Chandrakasan, A.~P.
\newblock Gazelle: A low latency framework for secure neural network inference.
\newblock \emph{{In} arXiv: 1801:05507}, 2018.

\bibitem[Kendall et~al.(2015)Kendall, Grimes, and Cipolla]{posnet}
Kendall, A., Grimes, M., and Cipolla, R.
\newblock Posnet: A convolutional network for real-time 6-dof camera
  relocalization.
\newblock \emph{In {Proceedings of the IEEE international conference on
  computer vision (ICCV)}}, 2015.

\bibitem[Krizhevsky \& Hinton(2009)Krizhevsky and Hinton]{cifar}
Krizhevsky, A. and Hinton, G.
\newblock Learning multiple layers of features from tiny images.
\newblock \emph{In {Computer Science Department, University of Toronto, Tech.
  Rep}}, 1, 2009.

\bibitem[Krizhevsky et~al.(2012)Krizhevsky, Sutskever, and Hinton]{alexnet}
Krizhevsky, A., Sutskever, I., and Hinton, G.~E.
\newblock Imagenet classification with deep convolutional neural networks.
\newblock \emph{In {NIPS}}, 2012.

\bibitem[LeCun et~al.(1998)LeCun, Bottou, Bengio, and Haffner]{lenet}
LeCun, Y., Bottou, L., Bengio, Y., and Haffner, P.
\newblock Gradient-based learning applied to document recognition.
\newblock \emph{In {Proceedings of the IEEE}}, 1998.

\bibitem[Li et~al.(2017)Li, Lai, Suda, Chandra, and Z.Pan]{PrivyNet}
Li, M., Lai, L., Suda, N., Chandra, V., and Z.Pan, D.
\newblock Privynet: A flexible framework for privacy-preserving deep neural
  network training with a fine-grained privacy control.
\newblock \emph{In {arXiv preprint arXiv:1709.06161}}, 2017.

\bibitem[Liu et~al.(2015)Liu, Luo, Wang, and Tang]{CelebA}
Liu, Z., Luo, P., Wang, X., and Tang, X.
\newblock Deep learning face attributes in the wild.
\newblock In \emph{Proceedings of International Conference on Computer Vision
  (ICCV)}, December 2015.

\bibitem[Mahendran \& Vedaldi(2015)Mahendran and
  Vedaldi]{mahendran2015understanding}
Mahendran, A. and Vedaldi, A.
\newblock Understanding deep image representations by inverting them.
\newblock \emph{Proceedings of the IEEE Conference on Computer Vision and
  Pattern Recognition (CVPR)}, pp.\  5188--5196, 2015.

\bibitem[Melis et~al.(2018)Melis, Song, De~Cristofaro, and
  Shmatikov]{melis2018exploiting}
Melis, L., Song, C., De~Cristofaro, E., and Shmatikov, V.
\newblock Exploiting unintended feature leakage in collaborative learning.
\newblock \emph{IEEE}, 2018.

\bibitem[Mohassel \& Zhang(2017)Mohassel and Zhang]{SecureML}
Mohassel, P. and Zhang, Y.
\newblock Secureml: A system for scalable privacy-preserving machine learning.
\newblock \emph{{In} 2017 38th IEEE Symposium on Security and Privacy (SP)},
  pp.\  19--38, 2017.

\bibitem[Osia et~al.(2017)Osia, Shamsabadi, Sajadmanesh, Taheri, Katevas,
  R.Rabiee, D.Lane, and Haddadi]{hybirddeeplearning}
Osia, S.~A., Shamsabadi, A.~S., Sajadmanesh, S., Taheri, A., Katevas, K.,
  R.Rabiee, H., D.Lane, N., and Haddadi, H.
\newblock A hybird deep learning architecture for privacy-preserving mobile
  analytics.
\newblock \emph{In {arXiv preprint arXiv:1703.02952}}, 2017.

\bibitem[Parcollet et~al.(2019)Parcollet, Ravanelli, Morchid, Linares,
  Trabelsi, Mori, and Bengio]{qlstm}
Parcollet, T., Ravanelli, M., Morchid, M., Linares, G., Trabelsi, C., Mori,
  R.~D., and Bengio, Y.
\newblock Quaternion recurrent neural networks.
\newblock \emph{In {International Conference on Learning Representations
  (ICLR)}}, 2019.

\bibitem[Reichert \& Serre(2014)Reichert and Serre]{Neuronal2014Reichert}
Reichert, D.~P. and Serre, T.
\newblock Neuronal synchrony in complex-valued deep networks.
\newblock \emph{{In} International Conference on Learning Representations
  (ICLR)}, 2014.

\bibitem[Ronneberger et~al.(2015)Ronneberger, Fischer, and Brox]{UNet}
Ronneberger, O., Fischer, P., and Brox, T.
\newblock U-net: Convolutional networks for biomedical image segmentation.
\newblock \emph{Proceedings of the IEEE Conference on Computer Vision and
  Pattern Recognition (CVPR)}, pp.\  234--241, 2015.

\bibitem[Shokri et~al.(2017)Shokri, Stronati, Song, and
  Shmatikov]{shokri2017membership}
Shokri, R., Stronati, M., Song, C., and Shmatikov, V.
\newblock Membership inference attacks against machine learning models.
\newblock \emph{IEEE Symposium on Security and Privacy (SP)}, pp.\  3--18,
  2017.

\bibitem[Simonyan \& Zisserman(2015)Simonyan and Zisserman]{vgg}
Simonyan, K. and Zisserman, A.
\newblock Very deep convolutional networks for large-scale image recognition.
\newblock \emph{In {ICLR}}, 2015.

\bibitem[Trabelsi et~al.(2018)Trabelsi, Bilaniuk, Zhang, Serdyuk, Subramanian,
  Santos, Mehri, Rostamzadeh, Bengio, and Pal]{Deep2018Trabelsi}
Trabelsi, C., Bilaniuk, O., Zhang, Y., Serdyuk, D., Subramanian, S., Santos,
  J.~F., Mehri, S., Rostamzadeh, N., Bengio, Y., and Pal, C.~J.
\newblock Deep complex networks.
\newblock \emph{In {International Conference on Learning Representations
  (ICLR)}}, 2018.

\bibitem[Wah et~al.(2011)Wah, Branson, Welinder, Perona, and Belongie]{CUB200}
Wah, C., Branson, S., Welinder, P., Perona, P., and Belongie, S.
\newblock {The Caltech-UCSD Birds-200-2011 Dataset}.
\newblock Technical Report CNS-TR-2011-001, California Institute of Technology,
  2011.

\bibitem[Wang et~al.(2018)Wang, Zhang, Bao, Zhu, Cao, and S.Yu]{notjustprivacy}
Wang, J., Zhang, J., Bao, W., Zhu, X., Cao, B., and S.Yu, P.
\newblock Not just privacy: Improving performance of private deep learning in
  mobile cloud.
\newblock \emph{In {KDD}}, 2018.

\bibitem[Xiang et~al.(2019)Xiang, Zhang, Ma, Zhang, Ren, and
  Zhang]{deepPrivacy}
Xiang, L., Zhang, H., Ma, H., Zhang, Y., Ren, J., and Zhang, Q.
\newblock Complex-valued neural networks for privacy protection.
\newblock In \emph{arXiv preprint arXiv:1901.09546}, 2019.

\bibitem[Zeiler \& Fergus(2014)Zeiler and Fergus]{zeiler2014visualizing}
Zeiler, M.~D. and Fergus, R.
\newblock Visualizing and understanding convolutional networks.
\newblock \emph{European Conference on Computer Vision (ECCV)}, pp.\  818--833,
  2014.

\bibitem[Zhang et~al.(2020)Zhang, Chen, Ma, Cheng, Ren, Xiang, Shi, and
  Zhang]{RENN}
Zhang, H., Chen, Y., Ma, H., Cheng, X., Ren, Q., Xiang, L., Shi, J., and Zhang,
  Q.
\newblock Rotation-equivariant neural networks for privacy protection.
\newblock \emph{In arXiv}, 2020.

\bibitem[Zhang et~al.(2016)Zhang, T.Yang, and Chen]{PrivacyZhang}
Zhang, Q., T.Yang, L., and Chen, Z.
\newblock Privacy preserving deep computation model on cloud for big data
  feature learning.
\newblock \emph{IEEE Transactions on Computers}, 65\penalty0 (5):\penalty0
  1351--1362, 2016.

\bibitem[Zhang et~al.(2019)Zhang, Chang, and Liang]{Yang2019Adversarial}
Zhang, Z., Chang, E.-C., and Liang, Z.
\newblock Adversarial neural network inversion via auxiliary knowledge
  alignment.
\newblock \emph{Proceedings of the Computer and Communications Security (CCS)},
  pp.\  225--240, 2019.

\bibitem[Zhu et~al.(2018)Zhu, Xu, Xu, and Chen]{qcnn}
Zhu, X., Xu, Y., Xu, H., and Chen, C.
\newblock Quaternion convolutional neural networks.
\newblock \emph{In {Proceedings of the European Conference on Computer Vision
  (ECCV)}}, pp.\  631--647, 2018.

\end{thebibliography}
\bibliographystyle{icml2020}

\end{document}